\documentclass{article}
\usepackage{ijcai21}

\usepackage{times}

\usepackage{soul}
\usepackage{url}
\usepackage[hidelinks]{hyperref}
\usepackage[utf8]{inputenc}
\usepackage[small]{caption}
\usepackage{graphicx}
\usepackage{amsmath}
\usepackage{booktabs}
\title{Relation Matters in Sampling: A Scalable Multi-Relational Graph Neural Network for Drug-Drug Interaction Prediction}

\author{
Arthur~Feeney$^{1}$\footnote{Equal Contribution}\and 
Rishabh~Gupta$^{1*}$\and 
Veronika~Thost$^{2}$\and
Rico~Angell$^1$\and\\
Gayathri~Chandu$^3$\and
Yash~Adhikari$^4$\and
Tengfei~Ma$^{2}$\footnote{Contact author: {tengfei.ma1@ibm.com}}
\affiliations
$^1$University of Massachusetts Amherst\\
$^2$IBM Research, MIT-IBM Watson AI Lab\\
$^3$Goldman Sachs\\
$^4$OPKO Diagnostics\\
\emails
}

\usepackage{amsmath}
\usepackage{amssymb}
\usepackage{xspace}
\usepackage{enumitem}
\usepackage{color, colortbl}
\usepackage{graphicx}
\usepackage{caption}
\usepackage{subcaption}
\usepackage{booktabs}

\newcommand{\deepddi}{\textsc{Drugbank}\xspace}
\newcommand{\twosides}{\textsc{Twosides100}\xspace}

\newcommand{\decagon}{{R-GAE}\xspace}
\newcommand{\ourmodelone}{{RW-GCN}\xspace}
\newcommand{\ourmodel}{{RS-GCN}\xspace}
\newcommand{\ifmodel}{{IFR-GAE}\xspace}

\usepackage{todonotes}
\usepackage{color, colortbl}

\begin{document}

\maketitle
\begin{abstract}
Sampling is an established technique to scale graph neural networks to large graphs. Current approaches however assume the graphs to be homogeneous in terms of relations and ignore relation types, critically important in biomedical graphs.
Multi-relational graphs contain various types of relations
that usually come with variable frequency and have different importance for the problem at hand. We propose an approach to modeling the importance of relation types for neighborhood sampling in graph neural networks and show that we can learn the right balance: relation-type probabilities that reflect both frequency and importance. Our experiments on drug-drug interaction prediction show that state-of-the-art graph neural networks profit from relation-dependent sampling in terms of both accuracy and efficiency.

\end{abstract}

\section{Introduction}
\label{sec:introduction}
Pharmacological interactions between drugs can cause serious adverse effects. Hence, predicting 
drug-drug-interactions (DDIs) is a task important for both drug development and medical practice: on the one hand, some diseases are best treated by combinations of drugs (e.g., antiviral drugs are typically administered as cocktails), on the other hand, one has to know about critical side effects\footnote{In a stricter sense, side effects are consequences of DDIs.} between new molecules and existing drugs. 
Considering the drugs as nodes and interactions as edges, DDI prediction can be considered as a link\footnote{We use the notions edge, relation, and link interchangeably.} prediction task in DDI graphs. 
There has been some recent progress based on deep learning tailored to DDI prediction \cite{ryu2018deep,caster19}, and especially graph neural networks (GNNs) show good performance \cite{zitnik2018modeling,xu2019mrgnn,genn19}.


A well-known challenge of applying GNNs is scalability. For example, in the graph convolutional network (GCN)~\cite{kip2017-GCN}, the number of nodes considered grows exponentially as the GCN goes deep. 
The special nature of biomedical graphs makes this problem even more serious:
In contrast to the kinds of sparse graphs in focus of AI research on link prediction (e.g., knowledge graphs (KGs) or citation networks), biomedical 
molecular interaction networks (e.g., containing drugs, proteins) are particularly dense; they also have large numbers of edges; various types of relations, many of which are undirected (i.e., symmetric); and often several 
relations between two nodes. 

In order to increase the efficiency of GNNs 
on large graphs, sampling methods have been proposed. However, these only focus on a general setting, that is, without considering the multi-relational nature of many graphs. Existing works include sampling random neighborhoods for individual nodes \cite{hamilton2017-GraphSAGE,chen2018stochastic,ying2018graph}, 
input for entire layers of the graph neural networks \cite{chen2018fastgcn,adaptive18}, or subgraphs as batches \cite{zeng2020graphsaint}. While some of these techniques are adaptive in that they learn to sample based on the problem at hand (vs. randomly) \cite{adaptive18,zeng2020graphsaint}, \emph{existing sampling methods 
usually focus on sampling nodes, while relation types are not regarded specifically. This may lead to unintended effects in the distribution of relations in the samples.} 

\begin{figure*}[ht]
 \centering
 \includegraphics[width=.7\textwidth]{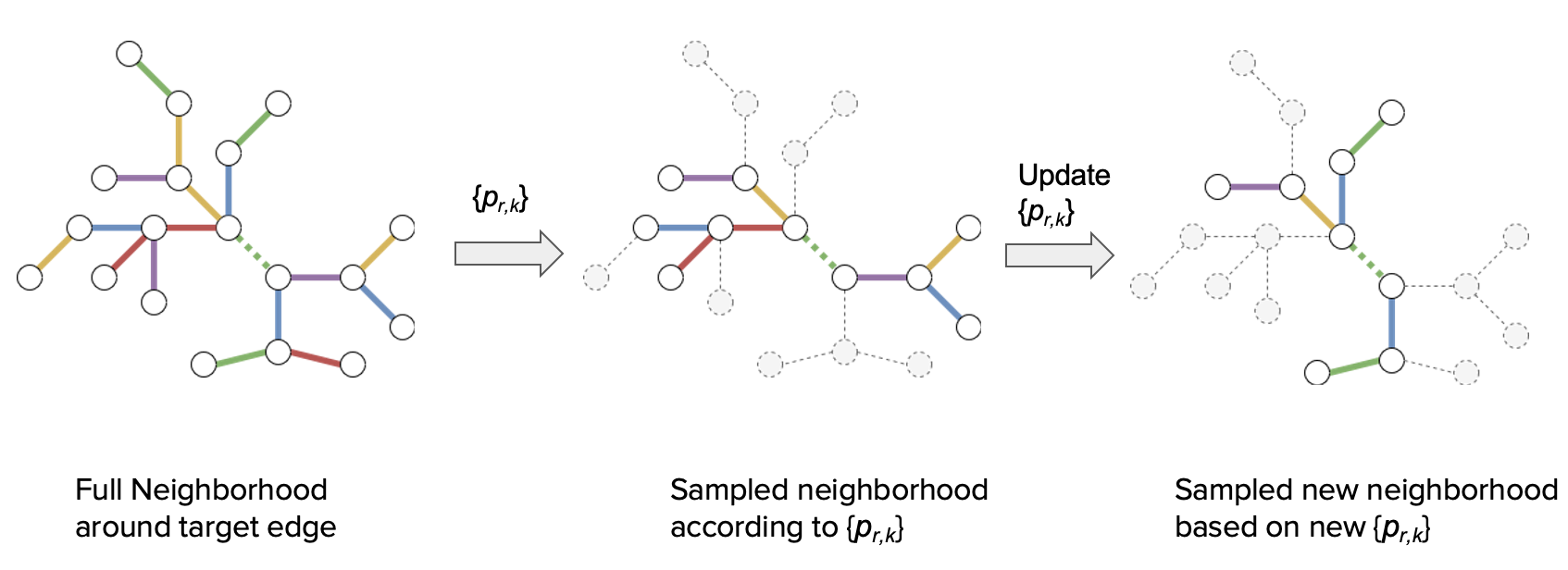}
 \caption{Visualization of the edge sampling in \ourmodel, which is based on 
 the sampling probability $p_{r,k}$ for edge type $r$ (depicted by different colors) in the $k^{th}$ hop. Since we focus on the link prediction task, we do sampling around both nodes in the given target edge~(dotted).}
\label{fig:sampling_process}
\end{figure*}

In this paper, \emph{we study how to conduct sampling in multi-relational graph neural networks for DDI prediction}. 
Specifically, we consider an extension of the relational graph convolutional network (R-GCN)~\cite{schlichtkrull2018modeling} and propose the \textbf{relation-dependent sampling-based graph convolutional network (RS-GCN)}. The core of our approach is to assign a learnable probability to each relation type and update it by a REINFORCE-based approach. The idea is outlined in Figure~\ref{fig:sampling_process}. We conducted experiments on two real-world DDI datasets and show that our models outperform state-of-the-art approaches in terms of both prediction performance and runtime efficiency. In addition, we investigate how the relation type impacts  the sampling. The results show that our model can learn the right balance: relation-type probabilities that reflect both frequency and importance, and also offer some kind of explanation.
%
%
In summary, our contributions are as follows:
\begin{itemize}
\item We propose a \emph{new sampling model for modeling relation-specific probabilities}  which particularly fits biomedical interaction networks (large, dense, heterogeneous). To the best of our knowledge, \ourmodel is the first model for sampling in multi-relational graphs.
\item We demonstrate that our model yields \emph{state-of-the-art performance for DDI prediction} on real-world datasets, while considerably improving efficiency.
\item We show that relation-specific sampling specifically \emph{benefits imbalanced data}, and give examples of how the learned relation probabilities can \emph{support explainability}.
\end{itemize}




\section{Related Work}
\label{sec:related}

\textbf{Drug-drug interaction prediction} models can be divided into three categories: (1) similarity-based approaches are based on the assumption that drug pairs have similar interaction patterns to drug pairs that are similar, and often compute concrete similarity metrics \cite{ddisimilarity,ma2018drug}; (2) structural representation learning computes an embedding for a drug by focusing on its molecule graph \cite{xu2019mrgnn}; (3) network relational learning focuses on the overall structure of the DDI network (though may include node and edge features) and essentially regards the task as link prediction \cite{zitnik2018modeling,genn19}.
Learning approaches are of course generally based on similarity, but models in the latter categories usually model the information they focus on more faithfully than by simple distance metrics. Our approach belongs to Category~(3). We do not intend to provide a full overview of the literature in DDI prediction since our focus is on the technical aspects of the problem in the context of graph neural networks. As mentioned above, GNNs have turned out as the type of AI technology best fitting biomedical graphs, due to their dense nature. 
\emph{Our work solves one challenge that is specific to the application of GNNs in the biomedical setting: dealing with graphs that are very dense and where edge types are critical.}


\textbf{Link prediction} approaches are actively studied for diverse applications and the 
variety of approaches is similarly large: tensor factorization \cite{fb15k}, heuristic models \cite{Lu2011LinkPI}, 
(bi)linear models \cite{yang2014embedding}, 
and more complex neural networks \cite{ryu2018deep} have been investigated in great number and depth.
Also several graph neural networks have been developed for that task and proven effective \cite{zitnik2018modeling,genn19,shang2019end,vashishth2020composition}. 
Most of these models work for multi-relational graphs, but little attention has been given to the edge types specifically so far.
Relational graph convolutional networks \cite{schlichtkrull2018modeling} learn one matrix of weight parameters per edge type, and have shown to outperform graph convolutional networks on multi-relational graphs. However, entire matrices can hardly be compared and hence serve less in terms of interpretability. 
In a similar vein, edge embeddings are considered in some graph neural networks \cite{vashishth2020composition,gilmer2017neural}. 
A single weight parameter per edge type is integrated into a GCN in \cite{shang2019end}. We will consider a similar approach as preliminary study to investigate how and what kind of edge-type \emph{probabilities} are learned within GCNs. 
So far, the challenges provided by biomedical networks have been mostly neglected in the very active research on link prediction, the most popular benchmark datasets are of very different nature. Our experiments will show that typical traditional and state-of-the-art link prediction approaches in KGs (non-GNNs) turn out disappointing for DDI prediction. Therefore, \emph{we focus on R-GCN and make it both more efficient and effective by using sampling based on edge-type probabilities.}

\textbf{Sampling for Graph Representation Learning}
has been extensively studied as a way to improve the computing time and memory use of GNNs, which become difficult to deploy as graph size increases. A commonly used method is GraphSAGE \cite{hamilton2017-GraphSAGE}, which extracts a random, fixed-size sample from the $k$-hop neighborhood around a target node. 
Layer-wise sampling approaches extract a fixed number of nodes per layer \cite{chen2018fastgcn,adaptive18}. Some of them are adaptive~\cite{adaptive18} in that they learn node-specific sampling probabilities but, to the best of our knowledge, edges have been not considered yet.
Most recent works have suggested the graph sampling method, which learn to sample entire subgraphs in the form of mini-batches \cite{zeng2020graphsaint} or as coarsened network input \cite{xu2020dynamically}. But the former only consider probabilities for nodes and edges, not for edge types, and the latter sample based on node-specific attention scores. In short, existing sampling approaches focus on dealing with large node numbers but do not consider the edge types.
\emph{We propose edge-type-specific neighborhood sampling (of edges) and show that it is beneficial in biomedical networks,  with might have many nodes too, but there the edge numbers are specifically challenging. } 



\section{Preliminaries}
\label{sec:preliminaries}

\paragraph{Multi-Relational graphs and DDI graphs.} Multi-relational graphs are of form $G = (V, E, \mathcal{R})$ with nodes (entities) $V$
and labeled edges (relations) $E\subseteq V\times \mathcal{R}\times V$, where $\mathcal{R}$ is a set of relation types. In this paper, we focus on DDI graphs containing drugs as nodes and various types of drug interactions as edges. 

\paragraph{DDI prediction.} DDI prediction can be regarded as link prediction task in a DDI graph. Given two drug nodes, the task is to predict whether there is an interaction between them and which interaction types we have for a given interaction (multi-label classification). 

\paragraph{Relational graph convolutional network.} Relational graph convolutional network (R-GCN) \cite{schlichtkrull2018modeling} is one of most commonly used multi-relational graph neural networks. It extends the classical graph convolutional network 
\cite{kip2017-GCN} to multi-relational graphs. GCNs update node embeddings by iteratively aggregating the embeddings of neighbor nodes.
R-GCN includes an additional weight matrix $W_r$ for each edge type $r$ and update the node embeddings in the $l^{th}$ layer as follows.
\begin{align}\label{eq:rgcn}
h_{u}^{l+1} = \phi\left(  
W_0^l h_u^l +
\sum_{r\in \mathcal{R}}
\sum_{v\in\mathcal{N}_{u,r}} 
{c_{u,r}}W_{r}^{l} h_{v}^{l} \right)
\end{align}
Here, $\mathcal{N}_{u,r}$ is the set of neighbors connected to node $u$ through edge type $r$; $c_{u,r}$ is a normalization constant, usually $\frac{1}{|\mathcal{N}_{u,r}|}$; 
$W_{r}^{l}$ are learnable weight matrices, one per $r\in \mathcal{R}$; 
and $\phi$ is a non-linear activation function. Obviously, in R-GCN, when the graph is dense and has a large number of relation types, the computational complexity of Equation~\eqref{eq:rgcn} is high. 
Our sampling approach is tailored to R-GCN, yet other multi-relational graph neural networks have the very same computational issues. An extension to those is left as future work.

\section{Modeling Edge-Type Probabilities} 
\label{sec:approach1}
Graph neural networks have proven successful in various tasks, such as for link prediction, and they have been successfully extended to multi-relational graphs. However, the computational complexity of these models is too high for very large graphs. Motivated by the solutions in homogeneous graphs, we propose a new sampling technique tailored to multi-relational graphs, based on probabilities for edge types. 

\subsection{Relation Sampling in Multi-Relational Message Passing}
\label{sec:modelone}
In some previous sampling approaches for GCNs~\cite{chen2018fastgcn,adaptive18}, the message-passing schema of GCN is rewritten to an expectation form over the prior distribution of all nodes. Reconsidering the formulation of R-GCN \eqref{eq:rgcn}, we can similarly reformulate it into an expectation form. However, instead of using the node distribution, we rather focus on the distribution of relations. When $c_{u,r}=\frac{1}{|\mathcal{N}_{u,r}|}$, the R-GCN Equation (\ref{eq:rgcn}) can be rewritten in the following expectation form.
\begin{align}
     h_u^{l+1} & = & \phi \left( W_0^l h_u^l  + |\mathcal{R}| \mathbb{E}_{p(r|u)} \mathbb{E}_{p(v|u,r)}( W_r^l h_v^l) \right) \notag\\
     & = & \phi \left( W_0^l h_u^l  + |\mathcal{R}| \mathbb{E}_{p(r,v|u)} ( W_r^l h_v^l) \right)\label{eq:expectation}
\end{align}
where  $p(r|u) = \frac{1}{|\mathcal{R}|}$ defines the probability of relation type $r$ conditioned on the node $u$ ($|\mathcal{R}|$ is the number of relation types, and the probability for each $r$ is the same); $p(v|u,r) = c_{u,r} = \frac{1}{|\mathcal{N}_{u,r}|}$ indicates a uniform distribution of neighbor nodes connecting to $u$ through relation $r$; and $p(r,v|u) = p(r|u)p(v|u,r)$ indicates the joint distribution of $r$ and $v$ given $u$, i.e., the probability of an edge $(u,r,v)$. 

To speed up message passing in a large, dense graph with many different relation types, a natural solution is to approximate the expectation component in Equation~\eqref{eq:expectation} by Monte Carlo sampling. For each node $u$, we can sample a set of $m$ edges in its neighborhood, approximating Equation~\eqref{eq:expectation}:
\begin{equation}
    h_u^{l+1} \simeq \phi \left( W_0^l h_u^l + \frac{|\mathcal{R}|}{m} \sum_{r,v \sim p(r,v|u)}  W_r^l h_v^l\right)
    \label{eq:rgcn_sampling}
\end{equation}

\subsection{\ourmodel: Relation Sampling based on Learned Probabilities}
\label{sec:probsampling}
In the R-GCN model, each relation type $r$ has the same node-dependent weight $c_{u,r}$ 
in the aggregation function, a simple normalization constant. However, we argue that the relation types matter. Different relation types occur with different frequency and have different semantics. Since these factors should largely impact the importance of relations, we posit that modeling relation-dependent probabilities is beneficial and propose to automatically learn them.

A straightforward idea to incorporate edge-type importance into R-GCN is to extend Equation~\eqref{eq:rgcn} by regarding $c_{u,r}$ as a learnable, normalized parameter. Specifically, we consider an additional, learnable parameter $l_r$ for each edge type $r\in\mathcal{R}$, and define the weight for each neighborhood as:
\begin{align}\label{eq:cur}
c_{u,r} &= \frac{e^{l_{r}}}{\sum_{r'\in\mathcal{R}} {|\mathcal{N}_{u,r'}|} e^{l_{r'}}}
\end{align}
We call this variant of R-GCN  \textbf{relation-weighted GCN (\ourmodelone)}. 
To obtain scalability, we could use random sampling with \ourmodelone. However, in such a setting, the sampling process is completely independent of edge types and the latter come only into play when computing the node embeddings, based on what was sampled randomly.

Therefore we develop the \textbf{relation-dependent sampling-based graph convolutional network (RS-GCN)} that \emph{learns} sampling probabilities for each edge type during training.

First, we derive an expectation form that is an alternative to the one from Equation~\eqref{eq:expectation}, by regarding a $c_{u,r}$ as in Equation~\eqref{eq:cur} and taking it as the probability $p(r,v|u)$ of an edge $(u,r,v)$ given $u$. Note that this is valid because, for a fixed $u$, $\sum_{r\in\mathcal{R},v \in\mathcal{N}_{u,r}} c_{u,r} =1$; and, similar to the original R-GCN in Equation~\eqref{eq:rgcn}, $c_{u,r}$ is the same for all neighbors $v$ for fixed $u$ and $r$. Thus Equation~(\ref{eq:rgcn}) can be rewritten as follows: 
\begin{eqnarray*}\label{eq:rgcn2}
h_{u}^{l+1} &=& \phi\left(  
W_0^l h_u^l +
\sum_{r\in\mathcal{R},v\in \mathcal{N}_{u,r}} 
p(v,r|u) W_{r}^{l} h_{v}^{l} \right)\\
&=& \phi \left( W_0^l h_u^l  + \mathbb{E}_{p(r,v|u)} ( W_r^l h_v^l) \right)
\end{eqnarray*}

Based on this expectation form, a message passing scheme for sampling $m$ nodes in the neighborhood of a node $u$ following the distribution of $p(r,v|u)$ is described by:
\begin{equation}
    h_u^{l+1} \simeq \phi \left( W_0^l h_u^l + \frac{1}{m} \sum_{r,v \sim p(r,v|u)}  W_r^l h_v^l\right)
    \label{eq:rs-gcn_sampling}
\end{equation}
where $p(r,v|u)$ equals the $c_{u,r}$ defined in Equation~(\ref{eq:cur}). As a last step, we will next refine the definition of $p(r,v|u)$.


The idea is to learn how to retain a useful neighborhood around an edge without compromising scalability, while tightly coupling message passing and sampling through the parameters. 
Similar to \ourmodelone, \ourmodel uses latent parameters $l_r$; they are initialized with elements $l_r \sim \mathcal{N}(0, 1)$. 
In each hop $k$ during sampling, we use the edge-type parameters $l_r$ to generate a distribution over all the edges in that hop and sample $n_k = s_k * |\mathcal{N}_{k-1}|$ edges 
with replacement; here $s_k$ is the number of edges to sample in hop $k$ for each node in hop $k-1$, $\mathcal{N}_{k-1}$ is the set of nodes sampled in hop $k-1$. 
The sampling probability for an edge with type $r$ in hop $k$ is:
\begin{align*}
    p_{r,k} &= \frac{e^{l_{r}}}{\sum_{r'\in\mathcal{R}} {|\mathcal{E}_{k,r'}|} e^{l_{r'}}}
\end{align*}
$\mathcal{E}_{k,r'}$ is the set of edges of relation type $r'$ in hop $k$.
This $p_{r,k}$ is used as $p(r,v|u)$ in Equation~\eqref{eq:rs-gcn_sampling}. 

We implement the sampling process using a method similar to GraphSAGE \cite{hamilton2017-GraphSAGE}. As shown in Figure~\ref{fig:sampling_process}, for each target edge (prediction), we sample a fixed-sized $k$-hop neighborhood for both nodes in it iteratively (i.e., for $k=1,2,\dots$), and then perform message passing using Equation~\eqref{eq:rs-gcn_sampling}. The main difference to GraphSAGE is that our method samples edges, not nodes, and that it \emph{learns} sampling probabilities for each relation type during training instead of using uniform sampling.


\subsection{Learning Relation Probabilities}
As there are difficulties in backpropagating through the sampled subgraph, we use the score function estimator REINFORCE \cite{reinforce}. For a loss function $L$ and sampled subgraph $g$, to estimate the gradient of the loss with respect to the latent parameters $l=(l_0,\dots,l_{|\mathcal{R}|})$, we compute $p_l(g)$, the probability of sampling $g$ given $l$:
\begin{align*}
    \nabla_{l} \mathop{{}\mathbb{E}_g} [L(g)] = \mathop{{}\mathbb{E}_g} [L(g) \nabla_{l} \log p_{l}(g)]
\end{align*}
In this way, REINFORCE does not require backpropagation through the message passing layers or the sampled subgraph if we can compute $\nabla_{l} \log p_{l}(g)$. We observe that this can be achieved easily by computing $\log p_{l}(g)$ as below and backpropagating the gradient; $r_i$ is the edge type of the $i^{th}$ edge sample: 
\begin{align*}
   \log p_{l}(g) &= \sum_{k} \sum_{i=1}^{n_k} \log p_{r_i,k} \\&
                 = \sum_{k} \sum_{i=1}^{n_k} \log {\frac{e^{l_{r_i}}}{\sum_{r'\in\mathcal{R}} {|\mathcal{E}_{k,r'}|} e^{l_{r'}}}}
\end{align*}
Note that REINFORCE gives an unbiased estimator but may incur high variance. The variance can be reduced by using control variate based variants of REINFORCE which is left as future work.

\paragraph{Discussion} Learning edge probabilities has been explored in different contexts. \cite{adaptive18} proposed a neural network to learn the optimum probability for importance sampling of the nodes in each layer. However, they focus on approximation of the homogeneous GCN model and their objective for optimum sampling is to reduce the variance. \cite{franceschi2019-LearningDiscreteStrucutes} incorporate a Bernoulli distribution for sampling the latent graph structures and, at the same time, automatically learn the hyperparameters of the distribution. In our case the graph structure is already given and our edge probability is assigned to known edges. Moreover, our sampling method is different form importance sampling~\cite{chen2018fastgcn}. For importance sampling, we have the fixed prior $p(r,v|u)$ and use another sampling probability $q(r,v)$ to approximate the expectation; but in our model we directly make $p(r,v|u)$ learnable and infer it from the model.

\section{Evaluation}
\label{sec:evaluation}

Out evaluation focuses on the following questions:

\textbf{{\large{Q1}}}
Does the incorporation of edge-type probabilities improve DDI prediction accuracy and efficiency?

\textbf{{\large{Q2}}}
Do learned probabilities offer improvement beyond fixed probabilities?

\textbf{{\large{Q3}}}
How do the learned probabilities distribute, and can we get any insights from them?


\subsection{Datasets}
We use the following two common DDI prediction datasets:

\textbf{\deepddi}\footnote{Not to be confused with the actual Drugbank database \cite{drugbank}; note that this name is also used in \cite{caster19}.}  \cite{ryu2018deep} contains drug-drug interactions extracted from Drugbank which can be either synergistic or antagonistic.
For 99.87\% of the drug pairs, there is only a single interaction type, thus we have nearly a multi-class classification problem (vs. multi-label).

\textbf{\textsc{Twosides}} \cite{Tatonetti2012DatadrivenPO} is a database of drug-drug interaction side effects. 73.27\% of the drug pairs have more than one type of side effect. In our experiments, we take a random sample of 100 side effect types and the corresponding data from \textsc{Twosides} (\twosides).  

Table~\ref{tab:datasets} shows statistics about the datasets. Also note that both of them have high average node degrees (\deepddi: 206.6, \twosides: 32.3), meaning they are very dense -- as it is typical for molecular interaction networks. 




\begin{table}[t]
\begin{center}
\begin{small}
\begin{tabular}{lcccr}
\toprule
&\# Nodes &
\# Edge Types & 
\# Edges\\
\midrule
\deepddi & 1,861&86 &192,284  \\
\twosides & 1,918 & 100 & 30,979 \\ 
\bottomrule
\end{tabular}%
\end{small}%
\end{center}%
\caption{Overview of datasets.}
\label{tab:datasets}
\end{table}

\subsection{Baselines}
We compare our model with state-of-the-art algorithms for DDI prediction as well as for knowledge graph completion.

\textbf{DistMult} \cite{yang2014embedding} is a common tensor factorization baseline for link prediction in knowledge graphs.

\textbf{DeepDDI} \cite{ryu2018deep} is a recent neural network for DDI prediction. 
It basically uses a feedforward neural network which takes the concatenation of two drug embeddings as input and predicts their interaction types.

\textbf{Message-passing neural network (MPNN)} \cite{gilmer2017neural} is a basic GNN architecture that uses message passing to update the node embeddings. The messages can include edge attributes (e.g., embeddings based on the edge types), thus it works well for multi-relational graphs. The final link prediction is done on node-pair embeddings.

\textbf{CompGCN} \cite{vashishth2020composition} is a state-of-the-art model for link prediction in KGs combining traditional, translation-based link prediction with GNN-based reasoning.

\textbf{\decagon} \cite{schlichtkrull2018modeling} is a relational graph autoencoder that has shown to perform well for DDI prediction \cite{zitnik2018modeling}. The encoder is an R-GCN for node embedding, and the decoder is a tensor factorization model which does link prediction based on the learned embeddings of a node pair. 

\textbf{\ifmodel} is a baseline in our ablation study. It differs from \decagon in that it applies a fixed sampling probability for each edge type based on the inverse frequency in the training data, assuming 
this balances type imbalance during sampling. 

For a fair comparison of the GNNs, we use random neighborhood sampling together with \decagon and \ourmodelone.  

\begin{figure*}[t]
     \centering
      \hfill
         \includegraphics[width=.4\textwidth]{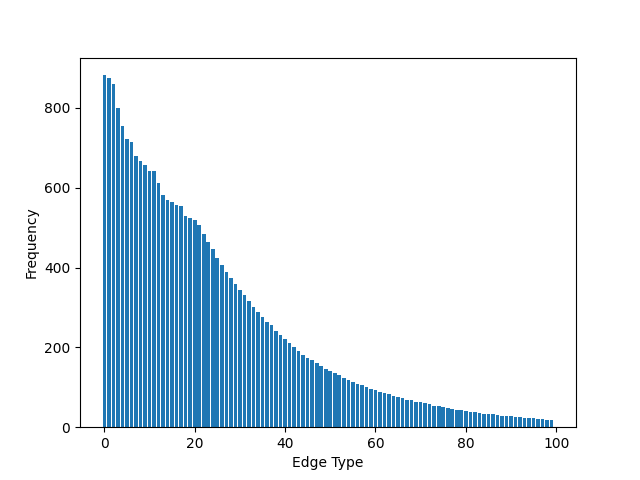}
    \hfill
         \includegraphics[width=.4\textwidth]{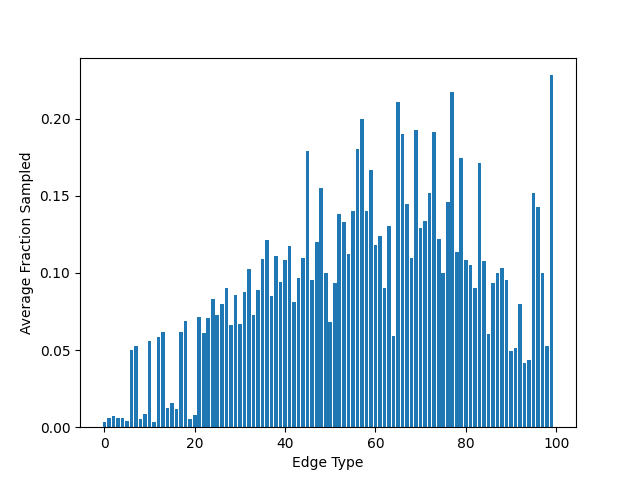}
     \hfill
     \caption{
     Distribution of edge types in the data (left) and 
     average fraction of edge types sampled in the evaluation (right) for \twosides.} 
\label{fig:pics}
\end{figure*}

\subsection{Model Configurations and Training}

We implemented our proposed approaches on top of \decagon. 
Specifically, we introduce learnable edge-type probabilities into the message-passing stage in the R-GCN encoder (\textbf{\ourmodelone}) and similar for the probabilistic sampling based on REINFORCE (\textbf{\ourmodel}) (see Section~\ref{sec:probsampling}). 
We implemented our models in PyTorch 
and PyTorch Geometric \cite{fey2019fast}. We use a standard 2-layer R-GCN with hidden dimension of size 128. 
%


We use similar settings for all models and datasets. For \deepddi we use a random 60/20/20 split as suggested in \cite{ryu2018deep}, and similar for \twosides. For \deepddi, we use the initial node features from \cite{ryu2018deep}, where the structural similarity profile 
of each drug is reduced in its dimension 
using PCA and then taken as node embedding. For \twosides, we use 
one-hot encodings. 

We process the target edges in batches using a batch size $b$ of 2000 for \deepddi and of 100 for \twosides. For each edge in the batch, we sample a fixed size $k$-hop neighborhood around that edge using the sampling procedure described in Section~\ref{sec:probsampling} with $k=2, s_1=7, s_2=3$. 
Additionally, during training, we sample random negative edges for each batch. The amount that we sample is equal to the batch size $b$. 
We use Adam optimizer with a learning rate of either $0.01$ or $0.001$, train with a patience of 100 epochs, and use binary cross entropy loss. 
\begin{table}[t]
\begin{center}
\begin{small}
\begin{tabular}{l|cc|cc} 
\toprule
&\multicolumn{2}{c|}{\deepddi}&\multicolumn{2}{c}{\twosides}\\
Model & PR-AUC & ROC-AUC& PR-AUC & ROC-AUC\\
\midrule
DistMult & 61.3 & 96.5 & 18.5 & 50.8    \\ 
DeepDDI & 75.0 & 98.7 & 28.6 & 60.7   \\
MPNN & 79.1 & 99.0 & 36.5 & 71.0    \\
CompGCN&84.3&98.7&34.0&67.6\\
\decagon & 80.6 & 98.9 & 35.3 & 67.1   \\
\midrule
\ifmodel & 81.5 & 98.9 & 37.3 & 71.3    \\
\textbf{\ourmodelone} & 82.8 & 99.2 & 35.7 & 68.0    \\
\textbf{\ourmodel} & \textbf{85.6} & \textbf{99.3} & \textbf{39.8} & \textbf{73.3}   \\
\bottomrule
\end{tabular}
\end{small}
\end{center}
\caption{DDI prediction results. } 
\label{tab:results}
\end{table}
\subsection{Results and Discussion}
We report our results in Table~\ref{tab:results}.
As common\cite{zitnik2018modeling}, we consider PR-AUC and ROC-AUC as metrics. 
PR-AUC captures the area under the plot of the precision rate against recall rate at various thresholds.
ROC-AUC quantifies the area under the plot of the true positive rate against the false positive rate at various thresholds.

We observe that GNN-based baselines outperform both MLP (DeepDDI) and the tensor factorization based models. This means that the information about the local neighborhood the GNNs encode is indeed useful. 
The generally lower numbers of \twosides can be explained by the fact that it is less dense that \deepddi and it also has a much larger percentage of drug pairs with more than one relation type. 
In particular, note that the state-of-the-art KG model CompGCN does not always outperform the rather basic R-GAE. That may be due to fact that the properties of DDI graphs are very different from the ones of typical knowledge graphs.

\subsubsection*{{\large{A1}}
Our incorporation of edge-type probabilities improves prediction accuracy and efficiency.}

From Table~\ref{tab:results}, we observe that \ourmodelone and \ourmodel perform overall better than all baselines on both datasets. This shows that giving different importance to different edge types based on the local neighborhood during message passing or neighbor sampling is useful in learning node representations in R-GCN. \ourmodel has an added advantage over \ourmodelone since it learns to ignore the less informative neighborhood during the sampling stage only and does not need to take less important edges into account during message passing.     

\begin{figure}
     \centering
         \includegraphics[width=0.42\textwidth]{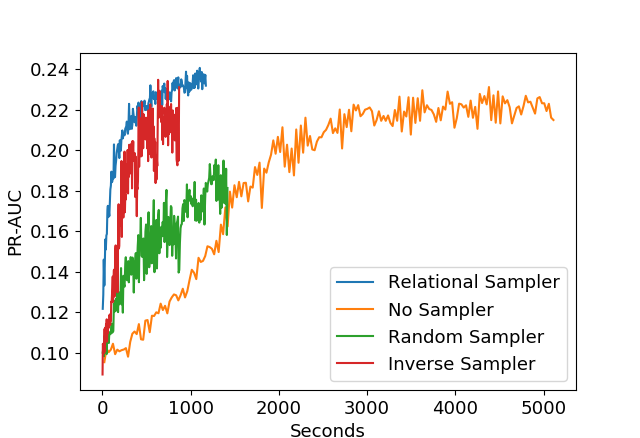}
        \caption{Runtime comparison of \ourmodel to \decagon with and without random sampling, on \twosides validation set. Our \ourmodel converges faster and reaches highest performance.}
        \label{fig:running_time}
\end{figure}

\ourmodel specifically provides an advantage in scalability as shown in Figure~\ref{fig:running_time}. 
Specifically, compared to standard \decagon (a version without sampling), we obtain an 
improvement during both training (2.47x) and inference at test time (1.92x); recall that \ourmodel did sampling in both training and inference. PR-AUC is not compromised but even better.
 
\subsubsection*{{\large{A2}}
Learned probabilities offer improvement beyond fixed probabilities based on inverse frequencies.} 

In order to show that it is indeed beneficial to \emph{learn} edge-type probabilities, we compare \ourmodel to \ifmodel. As it is shown in Table~\ref{tab:results}, on both datasets, \ifmodel outperforms the other baselines. This is likely due to the fact that both datasets are indeed imbalanced regarding edge types and shows that relation-dependent sampling is an effective means to deal with imbalanced data in general. The distribution of edge types in 
\twosides is depicted in Figure~\ref{fig:pics} (left). For \deepddi, see Appendix~\ref{app:deepddi}.

Nevertheless, \ifmodel does not achieve the performance of our models. One of the reasons behind this is that, even though \ifmodel takes class imbalance into account during sampling, the sampling probabilities are fixed globally and thus irrespective of the local neighborhoods. On the other hand, the sampling probabilities assigned to different edge types in \ourmodel are calculated adaptively based on the distribution of edge types in the local neighborhood of the target edge. We visualize the assigned probabilities in the $1$-hop neighborhood of a specific edge, whose edge type is to be predicted, in Appendix~\ref{app:neihborhood}. For demonstration purposes, we chose a neighborhood where all edge types occur exactly once. Especially for \twosides, we can see that, for that particular neighborhood, our model learns probabilities that are different from the ones of \ifmodel, since they do not reflect the inverse frequency of \twosides. 
Apart from capturing and resolving the imbalance of edge types in the data by assigning more weights to rare edge types, our model also learns to assign high weights to some more commonly occurring edge types based on their importance for the current task.
Hence, specific, learned probabilities generally reflect importance of edge types better. 
For \deepddi, the probabilities are similar to the ones learned by \ifmodel, which means that most edge types are probably of similar importance for this specific example task. 
However, since the performance of \ourmodel on \deepddi is better than the one of \ifmodel, the learned probabilities must capture some edge-type-specific information beyond frequency. 

Note that \ifmodel shows better performance than \ourmodelone on \twosides. Together with the fact that \ourmodel outperforms \ourmodelone, this confirms that dealing with edge-type probabilities during sampling is generally better than directly adding edge-type importance weights in the model. 

Figure \ref{fig:pics} (right) further underlines that our relation-specific sampling does not only solve imbalance in the data, but that it makes \ourmodel to better adapt to the link prediction tasks at hand. It shows the actual fractions of edge types sampled during inference averaged over all neighborhoods; hence, it depicts the learned probabilities on a global level. This averaging over the sampled neighborhoods of different target edges in the validation set confirms that, on  average, sampling for each type can be different from ``simple'' inverse frequencies depending on the dataset. Again, the corresponding figure for \deepddi is depicted in Appendix~\ref{app:deepddi}.



\subsubsection*{{\large{A3}} Observations.}
Overall, we observe that, for \deepddi, class imbalance has most influence on the learned probabilities. 
The most commonly occurring DDI (``increase in risk or severity of side effects'') is assigned the lowest probability, and the least occurring DDI (``increase in risk or severity of hyperkalemia'') is assigned the highest probability.
On the other hand, consider an example for \twosides, which has less imbalance. Here, the side effect ``traumatic haemorrhage'' that has almost the same frequency in our dataset as ``stress incontinence'' is assigned much higher probability ($0.06$) than latter ($5.6e^{-05}$). We  found several examples of this kind. 



\section{Conclusions}
\label{sec:conclusions}
Drug-drug-interaction graphs are large, dense and heterogeneous. In this paper, we propose a new relation-dependent sampling model to solve the scalability issue of graph neural networks on multi-relational graphs (such as DDI graphs). Our experiments on real-world DDI graphs show that \ourmodel outperforms state-of-the-art models 
in terms of both prediction performance and efficiency.


\bibliographystyle{named}
\bibliography{refs}

\begin{thebibliography}{}

\bibitem[\protect\citeauthoryear{Bordes \bgroup \em et al.\egroup
  }{2013}]{fb15k}
Antoine Bordes, Nicolas Usunier, Alberto Garcia-Duran, Jason Weston, and Oksana
  Yakhnenko.
\newblock Translating embeddings for modeling multi-relational data.
\newblock In {\em Proc. of NeurIPS}, pages 2787--2795. 2013.

\bibitem[\protect\citeauthoryear{Chen \bgroup \em et al.\egroup
  }{2018a}]{chen2018stochastic}
Jianfei Chen, Jun Zhu, and Le~Song.
\newblock Stochastic training of graph convolutional networks with variance
  reduction.
\newblock In {\em Proc. of ICML}, pages 942--950, 2018.

\bibitem[\protect\citeauthoryear{Chen \bgroup \em et al.\egroup
  }{2018b}]{chen2018fastgcn}
Jie Chen, Tengfei Ma, and Cao Xiao.
\newblock Fast{GCN}: Fast learning with graph convolutional networks via
  importance sampling.
\newblock In {\em Proc. of ICLR}, 2018.

\bibitem[\protect\citeauthoryear{DS \bgroup \em et al.\egroup
  }{2017}]{drugbank}
Wishart DS, Feunang YD, Guo AC, Lo~EJ, Marcu A, Grant JR, Sajed T, Li~C
  Johnson~D, Sayeeda Z, Assempour N, Iynkkaran I, Liu Y, Maciejewski A, Gale N,
  Wilson A, Chin L, Cummings R, Le~D, Pon A, Knox C, and Wilson M.
\newblock Drugbank 5.0: a major update to the drugbank database for 2018.
\newblock 2017.

\bibitem[\protect\citeauthoryear{Fey and Lenssen}{2019}]{fey2019fast}
Matthias Fey and Jan~Eric Lenssen.
\newblock Fast graph representation learning with pytorch geometric.
\newblock {\em arXiv preprint arXiv:1903.02428}, 2019.

\bibitem[\protect\citeauthoryear{Franceschi \bgroup \em et al.\egroup
  }{2019}]{franceschi2019-LearningDiscreteStrucutes}
Luca Franceschi, Mathias Niepert, Massimiliano Pontil, and Xiao He.
\newblock Learning discrete structures for graph neural networks.
\newblock In {\em Proc. of ICML}, 2019.

\bibitem[\protect\citeauthoryear{Gilmer \bgroup \em et al.\egroup
  }{2017}]{gilmer2017neural}
Justin Gilmer, Samuel~S Schoenholz, Patrick~F Riley, Oriol Vinyals, and
  George~E Dahl.
\newblock Neural message passing for quantum chemistry.
\newblock In {\em Proc. of ICML}, pages 1263--1272, 2017.

\bibitem[\protect\citeauthoryear{Hamilton \bgroup \em et al.\egroup
  }{2017}]{hamilton2017-GraphSAGE}
William Hamilton, Rex Ying, and Jure Leskovec.
\newblock Inductive representation learning on large graphs.
\newblock In {\em Proc. of NeurIPS}, 2017.

\bibitem[\protect\citeauthoryear{Huang \bgroup \em et al.\egroup
  }{2018}]{adaptive18}
Wenbing Huang, Tong Zhang, Yu~Rong, and Junzhou Huang.
\newblock Adaptive sampling towards fast graph representation learning.
\newblock In {\em Proc. of NeurIPS}, page 4563–4572, 2018.

\bibitem[\protect\citeauthoryear{Huang \bgroup \em et al.\egroup
  }{2019}]{caster19}
Kexin Huang, Cao Xiao, Trong~Nghia Hoang, Lucas~M. Glass, and Jimeng Sun.
\newblock {CASTER:} predicting drug interactions with chemical substructure
  representation.
\newblock {\em CoRR}, abs/1911.06446, 2019.

\bibitem[\protect\citeauthoryear{Kipf and Welling}{2017}]{kip2017-GCN}
Thomas Kipf and Max Welling.
\newblock Semi-supervised learning with graph convolutional neural networks.
\newblock In {\em Proc. of ICLR}, 2017.

\bibitem[\protect\citeauthoryear{Lu and Zhou}{2011}]{Lu2011LinkPI}
Linyuan Lu and Tao Zhou.
\newblock Link prediction in complex networks: A survey.
\newblock {\em ArXiv}, abs/1010.0725, 2011.

\bibitem[\protect\citeauthoryear{Ma \bgroup \em et al.\egroup
  }{2018}]{ma2018drug}
Tengfei Ma, Cao Xiao, Jiayu Zhou, and Fei Wang.
\newblock Drug similarity integration through attentive multi-view graph
  auto-encoders.
\newblock In {\em Proc. of IJCAI}, page 3477–3483, 2018.

\bibitem[\protect\citeauthoryear{Ma \bgroup \em et al.\egroup }{2019}]{genn19}
Tengfei Ma, Junyuan Shang, Cao Xiao, and Jimeng Sun.
\newblock {GENN:} predicting correlated drug-drug interactions with graph
  energy neural networks.
\newblock {\em CoRR}, abs/1910.02107, 2019.

\bibitem[\protect\citeauthoryear{Ryu \bgroup \em et al.\egroup
  }{2018}]{ryu2018deep}
Jae~Yong Ryu, Hyun~Uk Kim, and Sang~Yup Lee.
\newblock Deep learning improves prediction of drug--drug and drug--food
  interactions.
\newblock {\em Proc. of the National Academy of Sciences},
  115(18):E4304--E4311, 2018.

\bibitem[\protect\citeauthoryear{Schlichtkrull \bgroup \em et al.\egroup
  }{2018}]{schlichtkrull2018modeling}
Michael Schlichtkrull, Thomas~N Kipf, Peter Bloem, Rianne Van Den~Berg, Ivan
  Titov, and Max Welling.
\newblock Modeling relational data with graph convolutional networks.
\newblock In {\em Proc. of ESWC}, pages 593--607. Springer, 2018.

\bibitem[\protect\citeauthoryear{Shang \bgroup \em et al.\egroup
  }{2019}]{shang2019end}
Chao Shang, Yun Tang, Jing Huang, Jinbo Bi, Xiaodong He, and Bowen Zhou.
\newblock End-to-end structure-aware convolutional networks for knowledge base
  completion.
\newblock In {\em Proc. of AAAI}, pages 3060--3067, 2019.

\bibitem[\protect\citeauthoryear{Tatonetti \bgroup \em et al.\egroup
  }{2012}]{Tatonetti2012DatadrivenPO}
Nicholas~P. Tatonetti, Patrick Ye, Roxana Daneshjou, and Russ~B. Altman.
\newblock Data-driven prediction of drug effects and interactions.
\newblock {\em Science translational medicine}, 4 125:125ra31, 2012.

\bibitem[\protect\citeauthoryear{Vashishth \bgroup \em et al.\egroup
  }{2020}]{vashishth2020composition}
Shikhar Vashishth, Soumya Sanyal, Vikram Nitin, and Partha Talukdar.
\newblock Composition-based multi-relational graph convolutional networks.
\newblock {\em Proc. of ICLR}, 2020.

\bibitem[\protect\citeauthoryear{Vilar \bgroup \em et al.\egroup
  }{2014}]{ddisimilarity}
Santiago Vilar, Eugenio Uriarte, Lourdes Santana, Tal Lorberbaum, George
  Hripcsak, and Nicholas Tatonetti.
\newblock Similarity-based modeling in large-scale prediction of drug-drug
  interactions.
\newblock {\em Nature protocols}, 9:2147--2163, 09 2014.

\bibitem[\protect\citeauthoryear{Williams}{1992}]{reinforce}
Ronald~J. Williams.
\newblock Simple statistical gradient-following algorithms for connectionist
  reinforcement learning.
\newblock {\em Mach. Learn.}, 8(3–4):229–256, May 1992.

\bibitem[\protect\citeauthoryear{Xu \bgroup \em et al.\egroup
  }{2019}]{xu2019mrgnn}
Nuo Xu, Pinghui Wang, Long Chen, Jing Tao, and Junzhou Zhao.
\newblock {MR-GNN:} multi-resolution and dual graph neural network for
  predicting structured entity interactions.
\newblock {\em Proc. of IJCAI}, pages 3968--3974, 2019.

\bibitem[\protect\citeauthoryear{Xu \bgroup \em et al.\egroup
  }{2020}]{xu2020dynamically}
Xiaoran Xu, Wei Feng, Yunsheng Jiang, Xiaohui Xie, Zhiqing Sun, and Zhi-Hong
  Deng.
\newblock Dynamically pruned message passing networks for large-scale knowledge
  graph reasoning.
\newblock {\em Proc. of ICLR}, 2020.

\bibitem[\protect\citeauthoryear{Yang \bgroup \em et al.\egroup
  }{2015}]{yang2014embedding}
Bishan Yang, Wen{-}tau Yih, Xiaodong He, Jianfeng Gao, and Li~Deng.
\newblock Embedding entities and relations for learning and inference in
  knowledge bases.
\newblock {\em Proc. of ICLR}, 2015.

\bibitem[\protect\citeauthoryear{Ying \bgroup \em et al.\egroup
  }{2018}]{ying2018graph}
Rex Ying, Ruining He, Kaifeng Chen, Pong Eksombatchai, William~L Hamilton, and
  Jure Leskovec.
\newblock Graph convolutional neural networks for web-scale recommender
  systems.
\newblock In {\em Proc. of KDD}, pages 974--983, 2018.

\bibitem[\protect\citeauthoryear{Zeng \bgroup \em et al.\egroup
  }{2020}]{zeng2020graphsaint}
Hanqing Zeng, Hongkuan Zhou, Ajitesh Srivastava, Rajgopal Kannan, and Viktor
  Prasanna.
\newblock Graphsaint: Graph sampling based inductive learning method.
\newblock {\em Proc. of ICLR}, 2020.

\bibitem[\protect\citeauthoryear{Zitnik \bgroup \em et al.\egroup
  }{2018}]{zitnik2018modeling}
Marinka Zitnik, Monica Agrawal, and Jure Leskovec.
\newblock Modeling polypharmacy side effects with graph convolutional networks.
\newblock {\em Bioinformatics}, 34(13):i457--i466, 2018.

\end{thebibliography}

\appendix

\begin{figure*}[t]
     \centering
      \hfill
         \includegraphics[width=.42\textwidth]{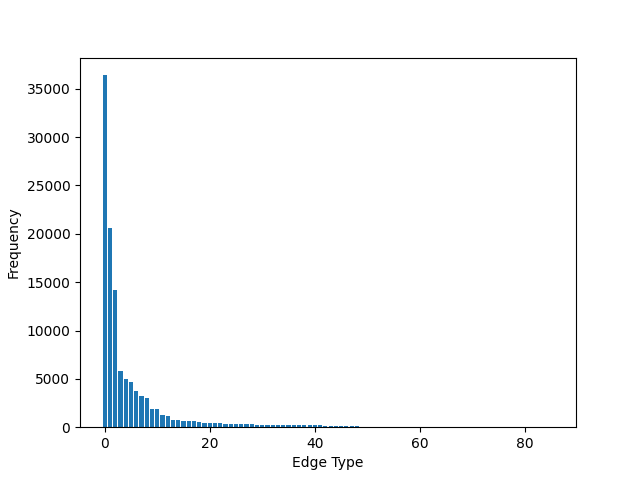}
    \hfill
         \includegraphics[width=.42\textwidth]{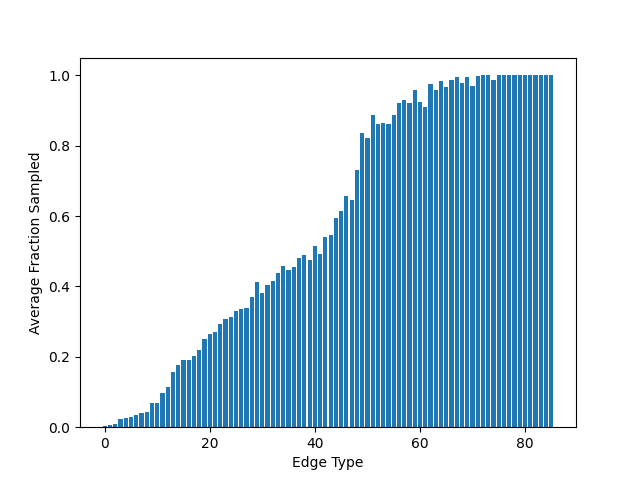}
     \hfill
     \caption{
     Distribution of edge types in the data (left) and 
     average fraction of edge types sampled in the evaluation (right) for \deepddi.} 
\label{fig:pics2}
\end{figure*}
\begin{figure*}[t]
     \centering
      \hfill
         \includegraphics[width=.42\textwidth]{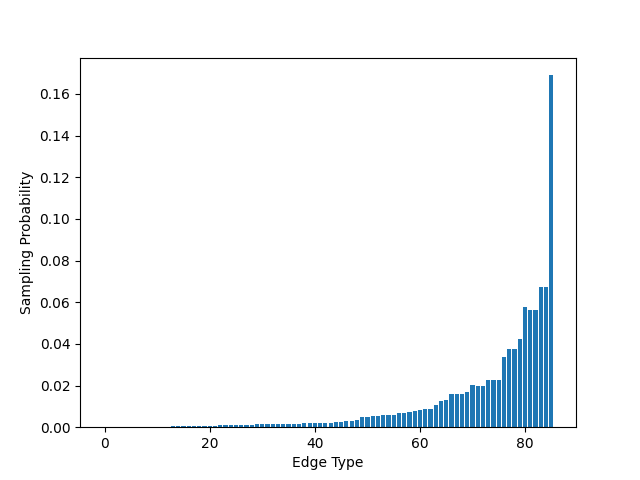}
     \hfill
         \includegraphics[width=.42\textwidth]{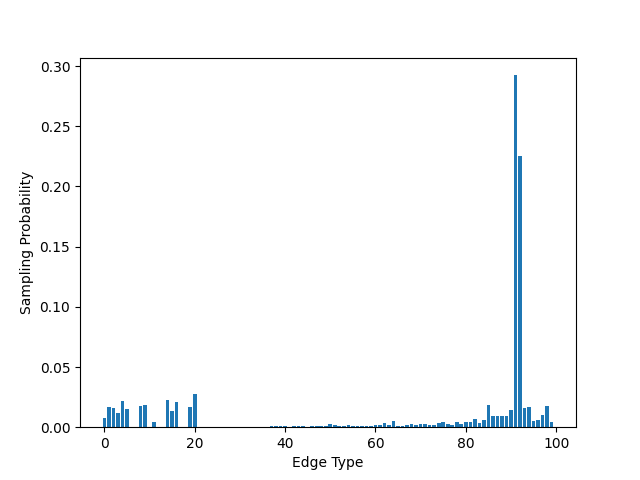}
     \hfill
     \caption{Learned probabilities for an example neighborhood for \deepddi (left) and \twosides (right).} 
\label{fig:pics3}
\end{figure*}

\section{Edge-Type Frequency and Learned Probabilities for \deepddi}\label{app:deepddi}

Figure~\ref{fig:pics2} depicts the distribution of edge types in \deepddi and our learned probabilities. The latter are very similar to the inverse frequencies, which means that sampling such that class imbalance is overcome seems to yield highest performance. Hence the edge types must be of similar importance.

\section{Additional Results}\label{app:neihborhood}

Figure~\ref{fig:pics3} depicts the probabilities learned for the edge types in an example neighborhood we sampled. While the ones for \deepddi closely reflect the inverse frequencies (compare to Figure~\ref{fig:pics2}), the ones for \twosides model different patterns of importance.

\end{document}


\maketitle

\appendix 

In the supplemental materials, we summarize the notations to clarify our method, and also add more experimental details, including the dataset creation, model architecture, and sampling details.

\section{Notations}
To clarify all the notations that we use in our method RS-GCN, we summarize them in Table~\ref{tab:notation}.

\begin{table}[ht]
\caption{Notations in the paper}
\label{tab:notation}
\vskip 0.10in
\begin{center}
\begin{small}
\begin{tabular}{ll}
\hline
$l_r$ & learnable parameter for edge type $r$ \\
$c_{u,r}$ & edge weight for edge type $r$ in message passing\\
$p_{u,r}$ & sampling probability for edge type $r$ around node $u$\\
$s_k$ & the fraction of sampled edges \\
$\mathcal{N}_{k-1}$ & set of nodes sampled in hop $k-1$ \\
$n_k$ & number of sampled edges \\
$g$ & sampled subgraph \\
$L$ & loss function \\
$p_l(g)$ & the probability of sampling a subgraph $g$ w.r.t. parameter $l$\\
\hline
\end{tabular}
\end{small}
\end{center}
\vskip -0.1in
\end{table}


\section{Additional Dataset Details}
\subsection{TWOSIDES100} 
To create TWOSIDES100, we take a subset of medium-occuring 100 side effects from the full TWOSIDES dataset \cite{Tatonetti2012DatadrivenPO}. We first sort the side effects by their frequency. We start from the 4,000th least common side effect and then consider every 60th side effect up to the 10,000th least common side effect.

\subsection{Negative Evidence} 
We extract mixture products from DrugBank \cite{drugbank} and synergism data from DrugCombDB \cite{drugcombdb}. When adding negative evidence to a dataset, we only include the negative edges that share an endpoint node with the target dataset. In the experiments on DrugBank DDI dataset, we only use the negative evidence from DrugBank because the synergism data implies that drug interaction is safe and not that there is no interaction.
For TWOSIDES, therefore, we use both the DrugBank and synergism datasets as negative evidence. In order to use the synergism data, we had to match drugs by name, rather than ID. This could have resulted in missing some possible data points.

\section{Model Architectures} 
When experimenting on DrugBank dataset, we always use a hidden dimension of 128. For TWOSIDES, we use a hidden dimension of 200. The specific architectures were the same for both datasets. For the MLP, we use three layers and apply batch normalization after each layer. For each of the GNN models, we use two graph layers followed by a classifier. For the MPNN, we use an MLP classifier and for the relational models, we use a tensor factorization method called dedicom similar to \cite{zitnik2018modeling}. Additionally for relational graph models, we use 30 bases for basis-decomposition \cite{schlichtkrull2018modeling}. We use a Binary Cross Entropy loss function for all models. We always use the ReLU activation function after each layer. Finally, we apply the sigmoid function to each model's output.

\section{Training Details}
We use a 60/20/20 train/validation/test split for both DrugBank and TWOSIDES. For DrugBank, we use the initial node features from \cite{ryu2018deep}, where the structural similarity profile (SSP) of each drug is reduced in its dimension (to 50) using PCA and then taken as node embedding. For TWOSIDES, we generate input features using one-hot encodings. We additionally perform batching on target edges: we create batches of edges in the input graph and sample a subgraph surrounding each of the target edges. We then use the sampled subgraph to make predictions for DDI types of the target edges. The batch size for DrugBank is 2000 and is 100 for TWOSIDES. We train each model for up to 300 epochs and use a patience of 100 epochs: we stop training when the PR-AUC has not increased for 100 epochs. All experiments use the ADAM optimizer with no weight decay and betas of 0.99 and 0.999. 
We use a learning rate of 0.001 in each experiment. 

\section{Sampling Details} 
Our sampling method is based on GraphSage and extracts a subgraph from the k-hop neighborhood surrounding target nodes \cite{hamilton2017-GraphSAGE}. As our experiments are restricted to networks with two GNN layers, we only sample from the two-hop neighborhood of the target nodes. For each method of sampling, we set an upper limit on the number of edges sampled in each hop. The upper limit in the first hop was set to be seven times the batch size. The limit for the second hop was three times the batch size. 

For all sampling methods, we compute logits for each edge and then apply a softmax function to each neighborhood to get probabilities for sampling the edges. For random sampling (in R-GAE), these logits are set to zero; each edge in a neighborhood has the same probability of being sampled. For the inverse frequency sampling (in IFR-GAE), we set the logits to be the log of the inverse of how many times the edge occurs. For the proposed learned sampling method (RS-GCN), we have parameters for each edge type that are summed depending on the input edge's types. Each sampling method uses the same underlying implementation. The only difference between them is how we compute the probabilities.

\bibliographystyle{plain}
\bibliography{refs}